# Transferability-based Chain Motion Mapping from Humans to Humanoids for Teleoperation

Matthew Stanley*, Yunsik Jung* , Michael Bowman, Lingfeng Tao, and Xiaoli Zhang

**Abstract**

Although data-driven motion mapping methods are promising to allow intuitive robot control and teleoperation that generate human-like robot movement, they normally require tedious pair-wise training for each specific human and robot pair. This paper proposes a transferability-based mapping scheme to allow new robot and human input systems to leverage the mapping of existing trained pairs to form a mapping transfer chain, which will reduce the number of new pair-specific mappings that need to be generated. The first part of the mapping schematic is the development of a Synergy Mapping via Dual-Autoencoder (SyDa) method. This method uses the latent features from two autoencoders to extract the common synergy of the two agents. Secondly, a transferability metric is created that approximates how well the mapping between a pair of agents will perform compared to another pair before creating the motion mapping models. Thus, it can guide the formation of an optimal mapping chain for the new human-robot pair. Experiments with human subjects and a Pepper robot demonstrated 1) The SyDa method improves the accuracy and generalizability of the pair mappings, 2) the SyDa method allows for bi-directional mapping that does not prioritize the direction of mapping motion, and 3) the transferability metric measures how compatible two agents are for accurate teleoperation. The combination of the SyDa method and transferability metric creates generalizable and accurate mapping need to create the transfer mapping chain.

## 1 Introduction

Teleoperation of a humanoid robot, in which a human operator remotely manipulates the robot through a control console, has the benefit of releasing humans performing dangerous task in unknown environments, such as space exploration, rescue operations, and healthcare service. Motion mapping techniques have been utilized for teleoperation as it allows the operator to use their body to intuitively control the humanoid robot. Good kinematic motion mapping and reconstruction are essential to generate human like and intricate robot movements [1,2]. Teleoperation tasks are difficult as the human operator does not have direct feedback from the environment. Therefore, the operator needs extensive training on controlling the robot and a good system for controlling the robot. This process can become very tedious if the controller does not accurate map the motion from the operator to the robot. This can become frustrating to the operator as it feels they do not have direct control over the robot. As such, motion mapping from the human operator to the robot resulting in intuitive control of the robot is in a great need [1,2].

Generating good data-driven motion mapping models can be difficult for multiple reasons. One of these challenges is the structural differences between humans and robots. A robot can have different degrees of freedom and/or different linkage lengths from the human trying to control it. Additionally, different human motion capturing approaches (such as using multiple cameras [3], attaching markers or sensors to the body [4,5,6], using wearable tracker suits [7,8], or using Kinect skeletal trackers [9,10,11,12]) used different coordinate systems and resolutions. These make it difficult to develop a one-to-one mapping between different human input systems and robot structures. Accordingly, most current data-driven motion mapping rules are system dependent, which means the data-driven mapping is only limited to the specific pair of human input and robot structure. Thus, a new mapping model must be created for each human and robot pair. To create data-driven models, individual models require pairwise data between a human input device and robot structure. Such pair-specific motion mapping cannot easily be adopted due to the burden of pair-specific data collection or the difficulty of mapping modeling for wide adoption of teleoperation.

Another open issue is the generalizability of the data-driven mapping methods. The mapping may only ensure good performance for the trained samples and can lead to degradation of the mapping performance for untrained data.

✉ Matthew Stanley
　mdstanley@mines.edu
　Yunsik Jung
　yunsikjung@mines.edu
　Michael Bowman
　mibowman@mines.edu
　Lingfeng Tao
　tao@mines.edu
　Xiaoli Zhang
　xlzhang@mines.edu

Matthew Stanley, Yunsik Jung, Michael Bowman, Lingfeng Tao, and Xiaoli Zhang are with Colorado School of Mines, Intelligent Robotics and System Lab, Golden, CO 80401 USA

Analytical methods developed the synergy-based motion mapping method, which defines the lower dimension information that describes the principal information of a human motion [13]. Since the motion of the robot should be the same motion for the operator, they should have the same synergy. The lower dimensional information extracted from the human are converted to the equivalent lower dimension information of the robot structure with an empirically defined synergy mapping rule [14,15,16]. Data-driven methods employed a similar way as synergy by using an autoencoder method, in which the latent features extracted by the autoencoder function like synergy-based lower dimension information [14]. While these methods reduce the complexity of the mapping tasks, the generalizability of these mapping is not ensured. The objective function of the mapping training is to minimize the error when using the latent features or the synergy parameters from human motion to robot motion. However, whether these features represent the true information of the human or the robot hand is not considered. In other words, the latent features can only ensure good mathematic mapping from human to robot for the training data rather than a generalizable set of features for other untrained motion trajectories/space.

To address the above two issues, we created what we refer to as a *transferability chain* method. The contributions of this method are as follows:
1. Develop a robust, accurate mapping schematic using a Synergy Mapping via Dual-Autoencoders (SyDa) model. This method aims to find similar synergies between the human's and the robot's motion. Along with the traditional losses for bot autoencoders, the SyDa extracts the synergy by making the latent features share with an extra loss function. This ensures that the mapping is more generalizable than using a traditional data-driven mapping method.
2. Introduce a chain motion mapping approach to generate mapping for a new human-robot pair (e.g., mapping the actions from a human tracking system to a specific robot) by chaining two existing intermediate pairs. This chain mapping approach allows for a reduction in required training for pair-specific models.
3. Define a transferability metric to guide the selection of existing intermediate pairs to minimize information loss and maximize the mapping performance. This transferability value is calculated by using mapping direction, structural discrepancy, and data noise that influence performance.
4. Evaluate the transferability-based motion mapping methods with a three-agent system including a human adult, a human kid, and a humanoid Pepper robot.

## 2 Related Works

### 2.1 Motion Mapping in Teleoperation

Conventional approaches for the kinematic motion mapping and reconstruction between humans and robots include empirically defined mapping rules and kinematics-model-based mapping. Various imitation trainings can determine the empirically defined mapping rules, for instance, humans imitating the motions of robots [1,2]. To achieve high accuracy, fine tuning of the empirically defined mapping rules is needed [4], which require high labor cost. To reduce the complexity of the mapping task, the synergy of the motion is studied [15,16,17]. Synergy defines the lower dimension information that describes parts of the given task [13]. While synergy-based approaches are often used, their reliant on linear reduction methods make them limited to use with more complex structures. The kinematic model-based approaches require kinematic models, (inverse) kinematic calculations, which cause high computational cost problems in real time processing [4,6,9,12].

### 2.2 Data-driven Mapping Methods

Data-driven techniques have an advantage over analytical approaches since they the high computation cost of controlling robots for imitating whole-body motion; however, it still has the limitation of the specific pairs. There are many approaches to using data-driven methods to map human motion onto a robot. [18] proposes a system that uses feed-forward neural networks to generate a robot's arm joint angles from the detected motions with multiple demonstrations for the model training. In addition, [1] implements a system that can map the motions to a humanoid robot by using feed-forward neural networks to calculate all joint angles. Kim et al. [19] used recurrent neural networks (RNN) to extract the features from the motions of a human and corresponding robot motions. To calculate the motions of the robot's joints, they use the RNN model which is trained by the back-propagation algorithm. [14] was able to extract synergy for mapping human motion to robot motion by using autoencoders to perform latent feature mapping. [20] uses convolutional autoencoders to generate avatar animations. All these mappings are structure dependent, limited to the specific pair that was used in the data collection and training process. The conventional data-driven mapping cannot easily be adopted due to this limitation.

## 3 Method

### 3.1 Synergy Mapping via Dual-Autoencoder

To improve the generalizability of a data-driven mapping model, the model should be able to leverage the synergy of the operator's motion to map to the robot. To obtain the synergy, we created a system of two autoencoders trained simultaneously called Synergy Mapping via Dual-Autoencoders (SyDa). The structure of the network and the training losses can be seen in Fig. 1 and can be used to map between two inputs. The SyDa model has two encoders and decoders, so the synergy from one source can be extracted and used to map it to either output. Consider the mapping of agents A and B with motion data defined as $\Theta_A = \{\theta_A \in WS_{M_A(\theta)}\}$ and $\Theta_B = \{\theta_B \in WS_{M_B(\theta)}\}$ where θ is the joint information and "WS" defines the reachable workspace that each agent can reach through their appendages. Each layer of the autoencoders uses this equation:

$$Z_i = \phi_i(a_i + \sum w_i * Z_{i-1}) \quad (1)$$

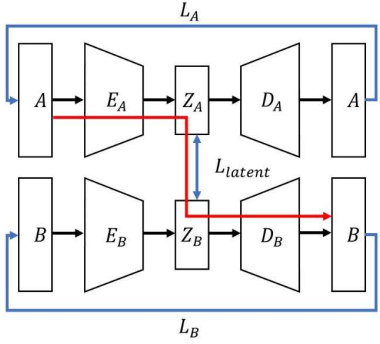

**Fig. 1** The SyDa method. The blue lines show the three loss functions. The red line shows how the input can be transferred.

In these equations, w, a, and $\phi$ are the weights, biases, and activation functions for each layer of the autoencoders. Considering agent A, the encoder layers begin with the input features ($Z_0 = \Theta_A$) and decrease in size at each step ($size(Z_i) < size(Z_{i-1})$), until the latent features reach the desired size ($Z_A$). After the desired latent feature size has been obtained, the decoder works in reverse, with $Z_0 = Z_A$, $size(Z_i) > size(Z_{i-1})$, and $Z_I = \Theta'_A$ where $\Theta'_A$ is the prediction of the input features. The same rules are applied to the autoencoder for agent B.

Using this structure for the two autoencoders in the SyDa method, the next step is to ensure that the latent features are properly being encoded into true synergies that cannot only map from human input to robot output but also can reconstruct to their corresponding inputs. This is achieved using three loss functions. The first two loss functions are between the input and the reconstructed output:

$$l_A = \|\Theta_A - \Theta'_A\| \quad (2)$$
$$l_B = \|\Theta_B - \Theta'_B\| \quad (3)$$

These two losses ensure that, after the joint information has been encoded then decoded, the information is still intact. While the latent features will represent the synergies of each agent, a third loss function is needed to ensure the synergies of two encoders are the same. By training these two autoencoders synchronously with pair-specific data, the third loss function can test that the latent features are the same:

$$l_{latent} = \|Z_A - Z_B\| \quad (4)$$

Since the latent features of each autoencoder is the same, it allows information to be encoded from either agent and be decoded into the form of the other. Since the latent features are trained to be the synergies that best represent the motion from each agent, less information is lost when mapping in either direction. This means motion can be from agent A to agent B and from agent B to agent A in a single model without one of the mappings becoming prioritized.

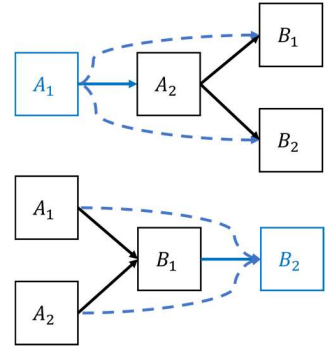

**Fig. 2** Examples of the chain method. The top figure shows that a new human input system ($A_1$) can be mapped to a pre-existing system ($A_2$) to control both robots ($B_1$ & $B_2$). Alternatively, the bottom figure shows how a new robot ($B_2$) can be mapped through an existing robot ($B_1$) that the human input systems ($A_1$ & $A_2$) can already control.

## 3.2 Chain Motion Mapping Across Structures

Using this SyDa model, multiple SyDa autoencoders can be chained in series to connect multiple models with each other. Consider a scenario with two human input systems ($A_i$) and two robots ($B_j$) where models exist to map motion between three of the four agents as shown in Fig. 2 (black arrows). The blue boxes are the fourth agent trying to interact with the others. Rather than training multiple mapping models such that new human systems are connected to each robot and new robots can be directly controlled by every human system (represented by the dashed blue line), a single mapping can be trained to connect the new agent with a similar agent (represented by the solid blue line). Using this chain transfer method, the actions of the new agent can be mapped to an agent (called intermediate agent, such as $A_2$ and $B_1$) that already has mappings to the other agents.

This benefit becomes more significant when more agents are involved. For a system with $n_h$ human motion capture systems trying to control any of $n_r$ different robots, the traditional approach would be to create a mapping for each type of user input with each type of robot output:

$$n_{models} = n_h \times n_r \quad (5)$$

With this new approach, significantly fewer models are required as a web can be generated such that every human-robot pair can be mapped through other agents. Since each agent must be connected to at least one other agent and have at least one path to all other agents, the minimal possible number of models required for this method is:

$$n_{models} \geq n_h + n_r - 1 \quad (6)$$

## 3.3 Transferability Estimation for Chain Optimization

While the chain transfer methods can reduce the number of new models than need to be trained, it does not guarantee the new pair will place optimally in the chain. To find the optimal transfer chain, a transferability metric is developed to determine which entities should be paired to maximize the performance of transferred mapping. The transferability metric considers how similar the two agents are for teleoperation along with the data collection noise. The assumption is that a pair of two agents with more similarities can more efficiently pass information between themselves (called transferability) and thus produce a better mapping performance. The higher the value that transferability is, the better mapping performance the pair will achieve. By knowing pair-specific transferability, it can be used to determine the optimal order for a chain.

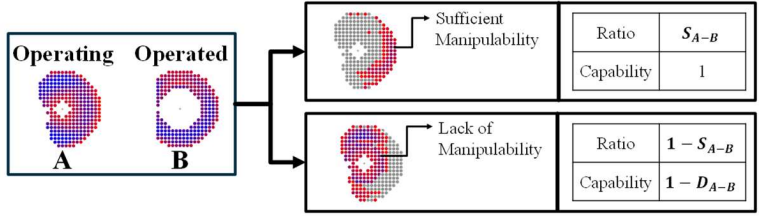

**Fig. 3** An illustration of the calculation transferability from $A$ to $B$ ($T_{A-B}$). The ratio of workspace for sufficient manipulability ($S_{A-B}$) and the dissimilarity ($D_{A-B}$) are considered for the transferability.

### 3.3.1 Definition of Transferability Metric

One of the key concepts of the transferability metric is to identify physical differences between operator $A$ controlling robot $B$. To address the size difference between the human and the humanoid robot, the length ratio ($L_{(A,B)}$) is defined as:

$$L_{(A,B)} = \frac{L_B}{max(L_A, L_B)} \quad (7)$$

Along with the length ratio, the manipulability metric is implemented to account for the difference in joint structures. For a given set of joint positions ($\theta$), the manipulability for a single agent is defined as $M(\theta) = \sqrt{det(JJ^T)}$, where $J$ is the Jacobian matrix [22,23]. As demonstrated in Fig. 3, comparing the manipulability metrics for the two agents result in two regions: the sufficient manipulability region where agent $B$ has more maneuverability ($WS_{(M_B-M_A \geq 0)}$), and the lack of manipulability region where agent $A$ has more maneuverability ($WS_{(M_B-M_A < 0)}$). To determine the manipulability, the relative size of each region and the difficulty of $B$ to imitate $A$ are needed. The size of the sufficient ratio is defined as:

$$S_{A-B} = \frac{WS_{(M_B-M_A \geq 0)}}{WS_{M_A}} \quad (8)$$

The ratio of the lack of manipulability region is thus defined as $1 - S_{A-B}$. In the sufficient manipulability region, $B$ should be able to reach any position that $A$ can, so the capability in this region is 1. For the lack of manipulability region, however, the difference needs to be accounted for. Using the Kullback Leibler Divergence [24], the dissimilarity ratio ($D$) in this region is defined in Equation 9. Thus, the capability in this region is ($1 - D_{A-B}$).

$$D_{A-B} = \sigma \left[ \sum M_A \, log \left( \frac{M_A}{M_B} \right) \right] \quad (9)$$

where, $\sigma$ is the sigmoid function to scale it down.

Since the ratio of the region depends on the size of the operator's workspace, the sufficient ratio and dissimilarity ratio will differ depending on which agent is the operator.

While the manipulability metric and length scale account for the physical differences of the agents, it does not consider the differences in the data collection accuracy. To normalize an influence of the noise in dataset, a coefficient factor, $\alpha$, is considered as the ratio of the noise of the sensor to the highest noise in the system. In the case of predicting the mapping performance, the value of α can be empirically estimated according to the type and resolution of data collection devices and processes of data collection and data analysis. Thus, the transferability from $A$ to $B$ can be evaluated as:

$$T_{A-B} = \alpha_{(A,B)} \times L_{(A,B)} \times [1 \times S_{A-B} + (1 - D_{A-B}) \times (1 - S_{A-B})] \quad (10)$$

### 3.3.2 Transferability-based Mapping Optimization

Knowing the pair-wise transferability, we can estimate the transferability when chaining models together. Consider a system of human systems $\{A_1,...,A_i\}$, robot subjects $\{B_1,...,B_j\}$, and a new robot subject $\{B_k\}$ (Fig. 4). Assuming that every human system is already mapped to the existing robots, the new robot only needs to connect with one existing robot in the web to build the mapping with every human system. The optimal existing robot to connect can be calculated by multiplying the transferability in each step:

$$T_{Optimal} = \underset{i,j}{argmax} \left( T_{A_i-B_j} \times T_{B_j-B_k} \right) \quad (11)$$

where $T_{A_i-B_j}$ is the transferability from $A_i$ to $B_j$ and $T_{B_j-B_k}$ is the transferability from $B_j$ to $B_k$. By comparing the transferability for every $i$ and $j$, one optimal mapping can be discovered instead of mapping every human system to the new robot. Similarly, this method can also support introducing new human inputs rather than a new robot.

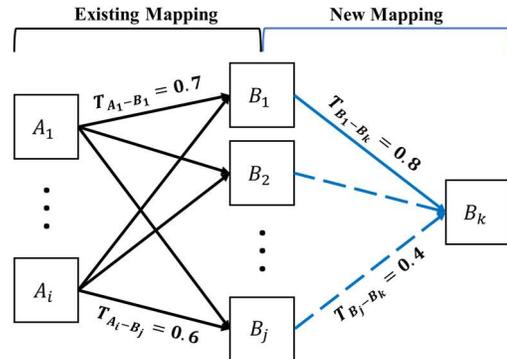

**Fig. 4** An illustration of transferability measures between existing mapping ($A_i$ to $B_j$) and new mapping ($B_j$ to $B_k$). The transferability of all the chains from $T_{A_1-B_1-B_k}$ to $T_{A_i-B_j-B_k}$ can be calculated, and the maximum transferability of all these chains will be chosen to build the mapping.

## 4 Experiments

### 4.1 Data Collection and Structures

To demonstrate the capability of the SyDa model in perform teleoperation, the method is tested in a teleoperation environment. The scenario was created where two human input systems are trying to control a humanoid Pepper robot. The human input systems are different in terms of coordinate systems (one origin at the head and the other at the torso) and size (a tall adult and a kid). To generate pairs of motion between the humans and Pepper, human participants were asked to follow and mimic the predefined Pepper motions (focusing on upper bodies to synchronize their motions). Having the humans mimic Pepper's motion ensures that the collected data contains points where the human's and Pepper's workspaces overlap. Using this method, 740 samples were collected for the pair between Pepper and adult, and 788 samples were collected for the pair between Pepper and kid. Further, the kid was asked to mimic the motion of the adult, where 660 samples were collected.

The Pepper consists of 19 joints and 19 degrees of freedom (DOFs). This joint information is presented as rotations in a unit quaternion with link lengths reported in meters. The human's skeleton motions recorded by the Microsoft Kinect has 15 joints and 31 DOFs. The joint information is presented as a point in cartesian space recorded in meters. Fig. 5 illustrates selected joints for the pepper and human. The two human subjects, despite using the same data collection method, have different origins with the adult's coordinate frame centered at the head and the kid's coordinate frame centered at the torso. The differences between each human operator and the Pepper robot can be seen in Table 1.

### 4.2 Mapping Evaluation

The performance of the SyDa model was compared against the performance of a direct autoencoder. The direct autoencoder uses the same encoding and decoding layers as the SyDa autoencoder to map A's motion as the input to output B's motion. Accordingly, the loss function for the direct autoencoder uses the L1-norm to match to the motion of A and B with the desired motion. Since the structure of the neural networks are the same and are trained using the same dataset, the difference between the two will mainly be due to the model structure and training method. To test against the full search space, 3-fold cross-validation was conducted to evaluate the accuracy of the model over the complete dataset. A magnitude of an average distance error, which is an average of distances between desired joint positions and their corresponding joint positions of the model prediction, is calculated to evaluate the performances of the training models.

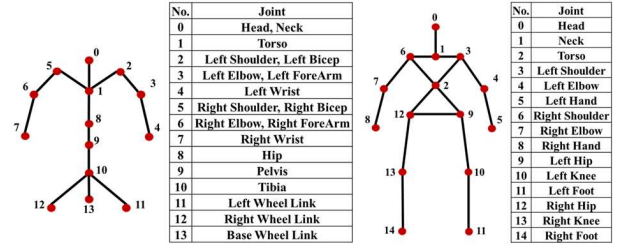

**Fig. 5** The Pepper (a) and human (b) skeletal structures with selected joints.

The next test is to evaluate the performance of the chain mapping method as predicted by the transferability metric. The mapping chain is created using both the SyDa autoencoder and the direct autoencoder, respectively, to verify the chain method performance. The models are created for both systems $H_2$-$H_1$-$R_1$ and $H_1$-$H_2$-$R_1$ to evaluate the relationship between the accuracy of the models with the transferability metric.

## 5 Results

### 5.1 Transferability Validation

The results from Table 2 indicate that the human to Pepper pairs (first two rows) are more accurate compared to the Pepper to human pairs (last two rows) with nearly double the average error. Even with the human to Pepper pairs and their corresponding Pepper to human pairs were trained with the same datasets (e.g., $H_2$-$R_1$ vs. $R_1$-$H_2$), the humans to Pepper pairs show better accurate mapping results than the corresponding Pepper to human pairs. This indicates that the direction of the mapping is significant for mapping quality. In addition, the average positional error (~10cm for each joint) is within the same order of magnitude as the reported accuracy for the Kinect [26]. This demonstrates that the errors from our models are within an acceptable order of magnitudes.

**TABLE 1**      SYMBOLS AND SPECIFICATIONS OF AGENTS: KID, ADULT, AND PEPPER

| Symbol | Agent | DOF | Height [mm] | Origin | Coordinate |
|---|---|---|---|---|---|
| $H_1$ | Kid | 31 | 1420 | Torso | Cartesian |
| $H_2$ | Adult | 31 | 1750 | Head | Cartesian |
| $R_1$ | Pepper | 19 | 1162 | Torso | Quaternion |

**TABLE 2**      RESULTS OF SUCCESS RATES THE TRANSFERABILITY ($T_{A\text{-}B}$) AND THE AVERAGE DISTANCE ERROR ($E_{A\text{-}B}$) OF THE MAPPING MODELS INCLUDING THE SUFFICIENT RATIO ($S_{A\text{-}B}$), THE CAPABILITY (1 - $D_{A\text{-}B}$) IN THE (1-$S_{A\text{-}B}$) REGION, AND THE COEFFICIENT FACTOR ($\alpha_{(A,B)}$)

| Mapping from A to B | $1 - S_{A-B}$ | $1 - D_{A-B}$ | $\alpha_{(A,B)}$ | $T_{A-B}$ | $E_{A-B}$ |
|---|---|---|---|---|---|
| $H_2$ to $R_1$ | 0.8878 | 0.3561 | 1 | 0.2358 | 0.0691 |
| $H_1$ to $R_1$ | 0.8884 | 0.3564 | 0.76 | 0.2335 | 0.0718 |
| $R_1$ to $H_1$ | 0.6738 | 0.0606 | 0.76 | 0.2002 | 0.1368 |
| $R_1$ to $H_2$ | 0.6781 | 0.0589 | 1 | 0.1991 | 0.1595 |

The transferability metric shows that the human to Pepper models have smaller similarity ratios and larger dissimilarity ratios compared to the Pepper to human models. Despite the larger value of ($1 - S_{A-B}$) in the humans to Pepper models, their relative lower dissimilarity in this region indicates human's motion can still be mapped to Pepper with a reasonable performance. Overall, the human to Pepper models have higher transferability values than the Pepper to human models, which confirmed the transferability is directional. To consider the influence of data quality, values of α are determined based on noise levels in the datasets.

Table 2 also validated the relationship between the performance of the mapping model and the transferability calculation. $H_2$-$R_1$ has the lowest distance error ($E_{H_2-R_1} = 0.0691m$) and highest transferability ($T_{H_2-R_1} = 0.2358$). Likewise, $R_1$-$H_2$ has the worst average distance error ($E_{R_1-H_2} = 0.1595m$) and the lowest transferability ($T_{R_1-H_2} = 0.1991$). An inverse relationship exists between the transferability and the reported error of the mapping. For the model which has the higher transferability, the lower mapping error can be achieved. If the pair has a higher transferability, it means that the workspace and manipulability of the output agent is more sufficient to mimic the input agent. Therefore, it is reasonable to conclude that the transferability metric is effective to estimate the performance rank of motion mapping after training.

## 5.2 Chain Mapping Across Structures

Table 3 shows the result for a chain system with $H_2$-$H_1$-$R_1$. The chain is created using a series of SyDa or a series of direct autoencoders, respectively. Table 3 shows that the accuracy increases by 9.3% when using the SyDa model instead of the direct autoencoder for the chain. Further, the result using the SyDa has the standard deviation decreased by 56.6% than the standard deviation using the direct autoencoder. These show the SyDa creates robust mapping with less variances than the direct autoencoder. This becomes significant for systems with more intermediate agents in the chain. As the chain grows, the small improvement from the SyDa model compounds and results in a better prediction at the end of the chain. Additionally, the results for the $H_2$-$H_1$-$R_1$ chain can be compared with the result of the $H_2$-$R_1$ model from Table 2. Comparing the SyDa model, the chain mapping method has a positional error, $E_{H_2-H_1-R_1}=0.0720m$, while the pair mapping model has an error, $E_{H_2-R_1} = 0.0691m$. While the chain does have a larger error and having an intermediate step in the chain does increase the model error, the error with a single intermediate step is less than the sum of the errors of the mapping models involved in this chain. The chain method produces an error that is within the same order of magnitude as the individual model.

An example of the chain transfer is shown in Fig. 6. As the figure suggests, some joint errors are compounding. While the Kinect returns human joints as distance from the relative center, Pepper joints are given as rotation from the previous joint. The chain mapping is agnostic to similar data type. The selected pairs used in a mapping chain can be in different coordinate systems or references frames, and the chain transfer method can solve different reference frames and coordinate systems on the fly and don't need standardized data before training. The chain with the SyDa model presents the output with higher accuracy than the chain with the direct autoencoder model. The error in Pepper's wrist position is a propagative function of all the rotational errors in Pepper's arm. The left hand demonstrates that the SyDa during the intermediate step more closely resembles the desired output compared to the direct autoencoder.

**TABLE 3**      THE AVERAGE DISTANCE ERRORS WITH THE STANDARD DEVIATIONS OF THE CHAIN $H_2$-$H_1$-$R_1$ USING THE SYDA AUTOENCODER AND THE DIRECT AUTOENCODER

|  | Right Elbow [$m$] | Right Wrist [$m$] | Left Elbow [$m$] | Left Wrist [$m$] | Total Average [$m$] |
|---|---|---|---|---|---|
| Real Pepper vs SyDa | $0.0609 \pm 0.0258$ | $0.0948 \pm 0.0501$ | $0.0482 \pm 0.0261$ | $0.0840 \pm 0.0536$ | $0.0720 \pm 0.0389$ |
| Real Pepper vs Direct Autoencoder | $0.0695 \pm 0.0641$ | $0.1124 \pm 0.0711$ | $0.0497 \pm 0.0369$ | $0.0832 \pm 0.0716$ | $0.0787 \pm 0.0609$ |

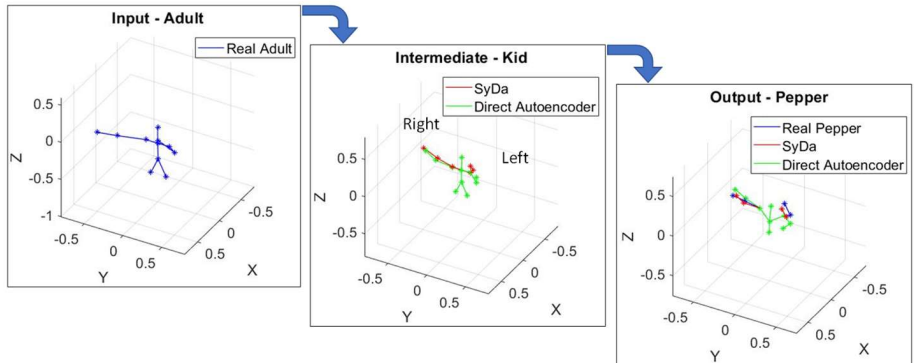

**Fig. 6** Performance of the trained Pepper using SyDa mapping. The left figure indicates the posture of adult as input, the middle one shows the intermediate predicted motions of kid, and the right one shows the predicted motions of Pepper as output. Comparisons for the left arm's postures shows that the SyDa (red) predicts closer to the desired value (blue) compared to the direct autoencoder (green).

## 5.3 Transferability-based Optimization for Mapping Chain

Table 4 shows the average distance error results when testing $H_1$-$H_2$-$R_1$. Compared to the other chain configuration in Table 3, the total average distance error increased by 58% for the SyDa model and 60% for the direct autoencoder model. In addition, the SyDa model showed a 10.8% increase in accuracy when compared to using the direct autoencoder model for the chain. It also has the standard deviation reduced by 25.1% than the standard deviation using the direct autoencoder. The error of the chain using the SyDa ($E_{H_1-H_2-R_1} = 0.1137m$) is higher than the error of the pair mapping model $H_1$-$R_1$ ($E_{H_1-R_1} = 0.0718m$), which can be expected as the same reason in Section 5.2.

The transferability measures calculated by Equation 11, and the average distance errors of those chain mappings using the SyDa autoencoder and the direct autoencoder are shown in Table 5. $T_{H_2-H_1-R_1}$ showed 23.6% increase in transferability measure than $T_{H_1-H_2-R_1}$. The relationships between the transferability measures and the average distance errors were found in the chain mappings. The higher the transferability of a mapping chain resulted in a lower error during the motion mapping. Thus, the transferability metric can predict the performances of chain mappings, and it holds true for using both SyDa and direct autoencoder in the chains, which further validate the effectiveness of transferability on mapping prediction. This consistent pattern would be a guidance to select a sequence which has a better performance for chain mappings with the same structures.

While the transferability does indicate which transfer chain will perform the best, it may not predict the accuracy (or error magnitude) of the transfer chain by itself. In Table 2, the transferability is similar while the Pepper to human errors are nearly double the human to Pepper errors. The transferability from Table 5 is nearly halved compared to those in Table 2, though the error is close to that of Table 2. This is because the mapping's accuracy is influenced by multiple factors including not only structure discrepancy and mapping direction, but also the quality of datasets. The purpose of the transferability study, therefore, is to rank the mapping performance of the chains with similar structures but not estimate the magnitude of the mapping accuracy.

Fig. 7 visualizes the two chains in the experiments. This system assumes that the mapping already exists between the two humans (black line), and the goal is to find the optimal placement for mapping to Pepper. The upper side of the figure (red) describes the motion transfer from $H_2$ through $H_1$ to $R_1$ while the lower side of the figure (blue) describes the motion transfer from $H_1$ through $H_2$ to $R_1$. The dashed lines indicate a path between two agents through an intermediate agent.

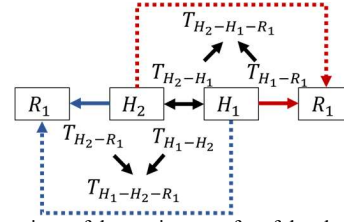

**Fig. 7** Configurations of the motion transfer of the chain adult to kid to Pepper ($H_2$-$H_1$-$R_1$ results in Table III) and the motion transfer of the chain kid to adult to Pepper ($H_1$-$H_2$-$R_1$ results in Table IV).

With the chain transfer method, $R_1$ only needs to pair one of two human systems to connect with both human systems. Choosing which entity $R_1$ pairs to may result in a different mapping performance, and the transferability measure can predict and optimize the mapping performance. For instance, consider pairing $R_1$ with $H_1$ (solid red arrow) first, then to find $H_2$ to $R_1$ through the chain mapping, the transferability measures for mapping $H_2$-$R_1$ (dashed red arrow) through the chain $H_2$-$H_1$-$R_1$ is $T_{H_2-H_1-R_1} = 0.1186$. This is greater performance than pairing $R_1$ with $H_2$ (solid blue line) first and measuring the transferability for $H_1$-$R_1$ (dashed blue line) through the chain $H_1$-$H_2$-$R_1$ which is $T_{H_1-H_2-R_1} = 0.1071$. Thus, the former chain will achieve higher mapping performance than the latter one. This transferability-based performance prediction is validated by actual errors of these mapping shown in Table V. Further details about the performances can be found in the supplementary video.

## 6 Discussion and Conclusion

The transferability mapping scheme presented in this paper is designed to reduce the training burden in systems with multiple human input systems and robots. The chain mapping scheme is developed to leverage pre-existing mappings to connect every human input system with every robot through other agents in the system. To achieve a more accurate mapping for the chain process, the SyDa is developed to produce a more generalized mapping across various agent models compared with traditional data-driven motion mapping methods. The experiment results demonstrate that the chain mapping's performance can be optimized by considering mapping direction, structure discrepancy, and data noise. Specifically, the results show that the mapping is directional, and it has an inverse correlation with the transferability metric. Using the transferability metric, agent pairs can be evaluated such that chains can be ranked by performance. The combination of these factors allows for system mapping with fewer trained models required compared to the traditional data-driven motion mapping approach. Through our three-agent system, it is demonstrated that the chain mapping scheme using the SyDa produces a good mapping across the chain.

**TABLE 4** THE AVERAGE DISTANCE ERRORS WITH THE STANDARD DEVIATIONS OF THE CHAIN $H_1$-$H_2$-$R_1$ USING THE SYDA AND THE DIRECT AUTOENCODER

|  | Right Elbow [m] | Right Wrist [m] | Left Elbow [m] | Left Wrist [m] | Total Average [m] |
|---|---|---|---|---|---|
| Real Pepper vs SyDa | $0.1040 \pm 0.0487$ | $0.1394 \pm 0.0603$ | $0.0704 \pm 0.0624$ | $0.1410 \pm 0.0837$ | $0.1137 \pm 0.0638$ |
| Real Pepper vs Direct Autoencoder | $0.1059 \pm 0.0559$ | $0.1817 \pm 0.0979$ | $0.0977 \pm 0.0762$ | $0.1189 \pm 0.0893$ | $0.1260 \pm 0.0798$ |

**TABLE 5** THE TRANSFERABILITY MEASURES AND THE AVERAGE DISTANCE ERRORS WITH THE STANDARD DEVIATIONS FOR THE CHAIN MAPPINGS

| Mapping | $T_{A-B-C}$ | $E_{A-B-C}$ (SyDa) [m] | $E_{A-B-C}$ (Direct Autoencoder) [m] |
|---|---|---|---|
| $H_2$ to $H_1$ to $R_1$ | 0.1186 | $0.0720 \pm 0.0389$ | $0.0787 \pm 0.0609$ |
| $H_1$ to $H_2$ to $R_1$ | 0.1071 | $0.1137 \pm 0.0638$ | $0.1260 \pm 0.0798$ |

This SyDa and chain transfer method well for generating mapping across three-agent-chain systems; however, the method does have some limitations. Since a small error occurs at each link of the chain, this error may aggregate across links. Therefore, a single chain cannot continue forever to include every robot and human operator. As the chain becomes longer, the system become more sensitive to small errors, which makes the resulting prediction less accurate. Thus, the transfer method will provide most benefits for small systems like three-agent chain scenarios, where a newly added agent needs to connect with existing pairs that have sufficiently good mapping. Note that it does not matter whether the existing pairs are trained directly or calculated through the prior transfer. If their mapping performance are sufficiently good, they can be added to the existing pairs and contribute to the new transfer behaviors.

# 7 Declaration

## 7.1 Funding

This material is based on work supported by the US NSF under grant 1652454. Any opinions, findings, and conclusions or recommendations expressed in this material are those of the authors and do not necessarily reflect those of the National Science Foundation.

## 7.2 Conflict of Interest

NA

## 7.3 Data and Material Availability

A copy of our dataset will be provided on the Intelligent-Robot-and-Systems-Laboratory Github page in the Synergy-DualAutoencoder repository. The Pepper motion data was collected using the Pepper ROS model. The human motion data was collected using the skeleton_marker package created by Patrick Goebel.

## 7.4 Code Availability

A copy of our code will be provided on the Intelligent-Robot-and-Systems-Laboratory Github page. The SyDa code can be found in the Synergy-DualAutoencoder repository. The code for the transferability analysis can be found in the Transferability repository. It was created using MATLAB without any toolboxes or applications.

## 7.5 Author's Contribution

All authors contributed to the study conception and design, such as initialization of the idea, review previous works, and material preparation. Matthew Stanley and Yunsik Jung performed the data collection, analysis, and visualization. The first draft of the manuscript was written by Xiaoli Zhang, Matthew Stanley, and Yunsik Jung. Michael Bowman and Lingfeng Tao provided comments and edits towards the creation of the final manuscript.

## 7.6 Ethics Approval

Ethical approval was waived by the local Ethics Committee of Colorado School of Mines in view of the retrospective nature of the study and all the procedures being performed were part of the routine care.

## 7.7 Consents to Participate

Informed consent was obtained from all individual participants included in the study.

## 7.8 Consents for Publication

The participants have consented to the submission of the case report to the journal.